\documentclass{article}

     \usepackage[nonatbib, final]{neurips_2020}

\usepackage[utf8]{inputenc} % allow utf-8 input
\usepackage[T1]{fontenc}    % use 8-bit T1 fonts
\usepackage{hyperref}       % hyperlinks
\usepackage{url}            % simple URL typesetting
\usepackage{booktabs}       % professional-quality tables
\usepackage{amsfonts}       % blackboard math symbols
\usepackage{nicefrac}       % compact symbols for 1/2, etc.
\usepackage{microtype}      % microtypography
\usepackage{algorithm}
\usepackage{algpseudocode}
\usepackage{amsmath}
\usepackage[english]{babel}
\usepackage{biblatex}
\usepackage{float}
\usepackage{graphicx, wrapfig, subcaption, setspace, booktabs}
\def\BibTeX{{\rm B\kern-.05em{\sc i\kern-.025em b}\kern-.08em
    T\kern-.1667em\lower.7ex\hbox{E}\kern-.125emX}}

\title{Reinforcement Learning for ConnectX}

\author{
  Shubham Gupta\\
  190020107\\
  Department of Mechanical Engineering\\
  IIT Bombay\\
  Mumbai\\
  \And
  Sheel Shah\\
  19D070052\\
  Department of Electrical Engineering\\
  IIT Bombay\\
  Mumbai\\
}

\begin{document}

\maketitle

\begin{abstract}
  ConnectX is a two-player game that generalizes the popular game Connect 4. The objective is to get X coins across a row, column, or diagonal of an M x N board. The first player to do so wins the game. The parameters (M, N, X) are allowed to change in each game, making ConnectX a novel and challenging problem. In this paper, we present our work on the implementation and modification of various reinforcement learning algorithms to play ConnectX.
\end{abstract}

\section{Introduction}
Connect 4 is a popular board game in which two players alternately take turns dropping their respective tokens on a $6 \times 7$ board by choosing the column they wish to drop their token in. The token falls down to the lowest unoccupied cell in the column, and if all six cells in a column are occupied, the move is declared illegal. The players’ objective is to place tokens so that there are 4 of their tokens consecutively across either a row or a column or a diagonal, and the first player to do so wins the game. Connect 4 has been solved \cite{connect4_solve_allis}, and it has been proved that the player who starts the game has a guaranteed-to-win strategy.

ConnectX is a generalization of Connect 4 that allows for variable board sizes and number of tokens to be connected (referred to as ’inarow’ hereafter). Hence, each time a player plays ConnectX, the board size and inarow may differ. This generalization makes the problem of optimally playing the game significantly more difficult. Guaranteed-to-win-or-draw strategies might exist for small board sizes and inarow lesser than 4, but no such strategy is known beyond these constraints. Therefore, a general method to play ConnectX is needed, and Reinforcement Learning is well-suited for the problem.

The following algorithms are implemented and modified to play ConnectX:
\begin{itemize}
    \item Greedy or One-Step Lookahead
    \item Monte Carlo Tree Search (MCTS)
    \item Minimax with alpha-beta pruning
    \item MCTS-minimax hybrid
    \item AlphaZero
\end{itemize}
MCTS-minimax hybrid is a novel algorithm which modifies MCTS to use informed rollouts, relying on minimax search, instead of random ones.

All these algorithms were implemented on an open-source ConnectX environment provided by \underline{\href{https://www.kaggle.com/competitions/connectx/overview}{Kaggle}}. The algorithms were played against one another and also submitted to a ConnectX competition hosted by Kaggle, where all submissions to the competition are matched by their position on the leaderboard, similar submissions played against each other, and the game's result used to update the leaderboard.

We were able to successfully implement the above mentioned algorithms, and their performances have been evaluated in section \ref{sec:results}. AlphaZero was the best performing algorithm, and we were able to achieve a leaderboard position of 9 (out 225 participating teams) with it.

\section{Related work}
To the best of our knowledge, there exists no previous work directly associated to playing ConnectX using Reinforcement Learning. On the other hand, Connect 4 has seen a lot of work, and serves as a good game to test algorithms. Connect 4 was first solved by James Dow Allen (October 1, 1988), and independently by Victor Allis (October 16, 1988) \cite{connect4_solve_allis}. Reference \cite{allen_2010} explores Connect 4 in great detail, and provides optimal strategies for both the players. Connect 4 has also been
solved in a brute-force manner by the following apporaches:
\begin{itemize}
    \item An 8-ply database \cite{tromp_8_ply}
    \item NegaMax / MiniMax using alpha-beta pruning
    \item NegaMax / MiniMax using transposition tables
\end{itemize}

Monte Carlo Tree Search \cite{mcts} (MCTS) is a popular framework that is well-applicable to problems where the environment can be simulated, but heuristics about good actions are unavailable. MCTS was combined with neural networks and the resulting algorithm AlphaGo \cite{alphago} performed extremely well in the board game of Go. This algorithm was then extended to other games whereby domain specific knowledge was removed from the algorithm to make it more general and thereby came the algorithm of AlphaZero, which we have adapted to ConnectX. The AlphaZero algorithm as described in \cite{alphazero} has been used to play various games like chess, go and shogi. However, extensions to similarly characterized games like Othello, Connect 4, and GoBang have also been made.

Some variants of MCTS \cite{mcts_minimax_heuristics} \cite{mcts_minimax_hybrid} incorporate minimax updates or heuristics to improve MCTS's performance. Variants that propagate proven wins/losses have been designed too and tested to be empirically better in games of sudden death \cite{mcts_solver}. MCTS has been adapted to play a plethora of games as in \cite{mcts_app1} \cite{mcts_app2} \cite{mcts_app3} \cite{mcts_app4} \cite{mcts_app5}, and has seen tremendous success.

\section{Defining the problem}
An instance of the game is parameterized by $(M, N, X)$, where $M$ is the number of rows in the board, $N$ the columns, and $X$ is the inarow parameter. A $player$ is given the current state of the board ($board$) and indication of the token he has to play ($mark$), and it must return a value $a \in \{0, ..., N-1\}$ denoting the column it wants to place the token $mark$ in. If the game ends (due to a win/draw/illegal move for $mark$), the simulation is stopped. If not, the opponent is provided the new $board$ and $mark$ for choosing a move. In the Kaggle environment, each $player$ has 5 seconds to choose their move, and if the $player$ fails to do so, it is automatically declared as the loser. An illegal choice of move also results in an immediate loss.

We hence want to design a function $player: board \times mark \rightarrow \{0, 1, ..., N-1\}$ that attempts to be the first player to get $X$ in a row on a board of size $M \times N$.

\section{Methodology}
% We begin with a description of the algorithms we have implemented.
\subsection{The greedy algorithm}
Algorithm \ref{alg:greedy} was implemented to act as a baseline, and to get a better idea of the functioning of the Kaggle environment and submissions.
\begin{algorithm}[h!]
\caption{Greedy Algorithm}\label{alg:greedy}
\begin{algorithmic}
\Require $board, mark$
\State $value_i \gets 0\ \forall i \in \{0, 1, ..., N-1\}$
\For{$i \in \{0, 1, ..., N-1\}$}
    \If{placing $mark$ in column $i$ connects $X$ in a row for $mark$}
    \State $value_i \gets \infty$
    \EndIf
    \State return $a = arg \max_i value_i$
\EndFor
\end{algorithmic}
\end{algorithm}

\subsection{Plain Monte Carlo Tree Search}
Each round of MCTS uses the following steps in order:
\begin{itemize}
    \item \textbf{Selection:} Starting from the current state, the move (and hence the next state) is chosen via a Upper-Confidence-Bound-like method that represents values of sub-trees. The exact formulation is:
    \begin{align*}
        UCB(\text{Tree rooted at state }S) &= \frac{w_S}{n_S} + c \cdot \sqrt{\frac{ln\ N_S}{n_S}}\\
        where:\\
        w_S &= \text{The number of winning simulations from state }S\\
        n_S &= \text{The total number of simulations from state }S\\
        N_s  &= \text{The total number of simulations from the parent of state }S\\
        c &= \sqrt{2}\ (\text{in our implementation})\\
    \end{align*}
    At each selection step, MCTS chooses the child state with the highest UCB(tree) value. This selection is repeated until a state that has atleast one unsimulated child is reached.
    
    \item \textbf{Expansion:} If the state at the end of selection is a terminal state, we skip to the backpropagation step. Otherwise, we pick a random valid move, and initialize a tree node representing the new state seen.
    \item \textbf{Simulation:} The new state seen is simulated. In plain MCTS, random legal moves are played till the game ends, and the result of this random playout is then backpropagated.
    \item \textbf{Backpropagation:} The path from the expanded state to the root state is traced, and the $w$ and $n$ for each of these states is updated based on whether the simulation was a win or a loss. A draw is treated as half a win.
\end{itemize}

\subsection{Minimax}
Algorithm \ref{alg:minimax} is a standard alpha-beta pruned minimax algorithm that was implemented. A search depth of 5 was chosen in accordance with 5 second move timeout. The heuristic used was as follows:\\

\begin{align*}
   heuristic = &\sum_{i=2}^X 1000^{i-1} \cdot \texttt{\# occurrences of i mark-tokens in a row}\ - \\
   &\sum_{i=2}^X 2000^{i-1} \cdot \texttt{\# occurrences of i opponent-mark-tokens in a row}
\end{align*}\\
Here, 1000 and 2000 are empirically chosen constants.

\begin{algorithm}[h!]
\caption{Alpha-Beta pruned minimax algorithm}\label{alg:minimax}
\begin{algorithmic}
\Require $board, mark, heuristic$
\Function{minimax}{state, depth, $\alpha$, $\beta$, maximizingPlayer}
    \If{depth = 0 or game ended}
    \State \Return $heuristic(state, mark)$
    \EndIf
    \If{maximizingPlayer}
    \State $value \gets -\infty$
    \For{each legal move}
    \State state' $\gets$ state reached by playing this move
    \State $value \gets max(value, MINIMAX(state', depth-1, \alpha, \beta, FALSE)$
    \If{$value \geq \beta$}
    \State \textbf{break}
    \EndIf
    \State $\alpha = max(\alpha, value)$
    \EndFor
    \Else
    \State $value \gets \infty$
    \For{each legal move}
    \State state' $\gets$ state reached by playing this move
    \State $value \gets min(value, MINIMAX(state', depth-1, \alpha, \beta, TRUE)$
    \If{$value \leq \alpha$}
    \State \textbf{break}
    \EndIf
    \State $\beta = min(\beta, value)$
    \EndFor
    \EndIf
\EndFunction
\State MINIMAX(board, 5, $-\infty$, $\infty$, TRUE)
\end{algorithmic}
\end{algorithm}

\subsection{MCTS with minimax rollouts}
We implemented a variation of plain MCTS by changing the \textbf{Simulation} step. Shallow minimax searches were used to guide the rollouts instead of random moves. A search depth of 2 was used, and all other steps were identical to that of plain MCTS. Since minimax rollouts slowed down each simulation by a significant amount, the number of iterations the algorithm could run per move dropped and hence, $c$ for the calculation of UCB was reduced to $\frac{1}{\sqrt{2}}$. A smaller $c$ was seen to empirically improve performance, and it can be argued having a smaller horizon calls for lesser (and better) exploration (and hence a smaller $c$).
\subsection{AlphaZero}

\begin{figure}[h!]
  \centering
  \includegraphics[height = 70mm]{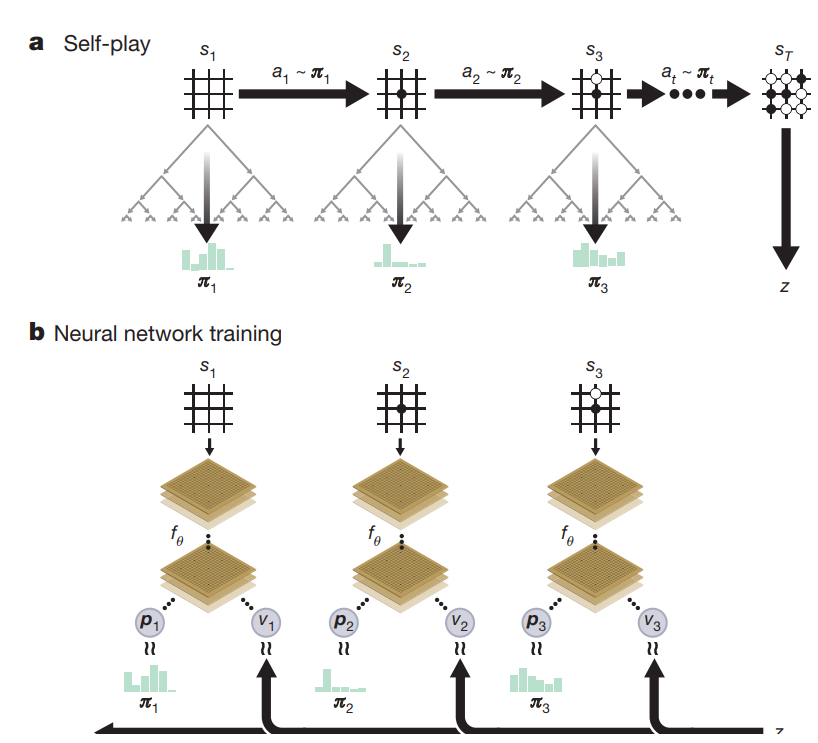}
  \caption{Self Play reinforcement learning in AlphaZero taken from \cite{Silver2017} }
  \label{sds_wait}
\end{figure}

The AlphaZero algorithm uses MCTS to initialize a search on the game tree and generates data on the game. This is called as the self play stage. At each step the action chosen is the one with the highest UCB value. This generates a probability vector $\pi$ over the moves. After the terminal state is reached the game is scored according to the rules and a win/loss/draw is declared. There is a concatenated neural network which maps the current state $s_t$ to both, the policy $\textbf{p}$ and the value function $v_t$. The neural network parameters are then updated using backpropagation, with an Adam optimiser, to minimise the error in the value $v_t$ using MSE loss as well as the difference between the policy vector $\textbf{p}$ and the search probabilities $\pi$ using cross entropy loss. Experience replay is also used here since it allows us to work with fewer simulations for each iteration since the training data keeps growing with iterations.

\begin{algorithm}[h!]
\caption{Policy Iteration through Self Play}\label{alg:policyiteration}
\begin{algorithmic}
\Require $executeEpisode,\ trainNN,\ initNN,\ threshold$
\Function{POLICY\_ITERATION}{numIters, numEpisodes}
    \State $\theta \gets initNN()$
    \State $trainExamples \gets []$
    \For{$ i \; in \{1,...,numIters\} $}
        \For{$ e \; in \{1,...,numEpisodes\} $}
            \State $ex \gets executeEpisode(nn)$
            \State $trainExamples.append(ex)$
        \EndFor
        \State $\theta_{new} \gets trainNN(trainExampples)$
        \If{percentage($\theta_{new}$ beats $\theta$) $\geq threshold$}
        \State $\theta \gets \theta_{new}$
        \EndIf
    \EndFor
\EndFunction
\end{algorithmic}
\end{algorithm}

\begin{algorithm}[h!]
\caption{Execute Episode}\label{alg:execep}
\begin{algorithmic}
\Require $assignReward, MCTS, gameNextState, gameStartState$
\Function{executeEpisode}{numSims}
    \State $examples \gets []$
    \State $s \gets gameStartState()$
    \While {True}
        \For{$i \; in [1,...,numSims]$}
            \State $MCTS(s,\theta)$
        \EndFor
        \State $examples.add((s,\pi_s,\_))$
        \State $a* \sim \pi_s$
        \State $s \gets gameNextState(s,a*)$
        \If{$gameEnded(s)$}
            \State $//{Fill\; \_ \; with\; rewards\; in\; examples, \;the \;backpropagation\; step\; for\; MCTS}$
            \State $examples \gets assignReward(examples)$
            \State \textbf{break and} \Return $examples$
        \EndIf
    \EndWhile
\EndFunction
\end{algorithmic}
\end{algorithm}

After training, the new network is played against the older network before the current round of training. And if the new network performs sufficiently better compared to the previous one,i.e., out of the total decided games which were played (not draw) if the fraction is decided games won by newer network is above a certain threshold (0.6 in our case), the network in use is updated and otherwise the newer network is not used.

Since ConnectX does not fix the size of the board beings used it poses a problem since the input state size is not always fixed. The workaround we found for that is using padding to make the input size always as $12\times12$ so that the state is always fully represented. The number was chosen at random based on how the board sizes varied during the competition.

\section{Results}

\label{sec:results}
\begin{table}[H]
  \caption{Comparison of agents}
  \label{comp_table}
  \centering
  \begin{tabular}{ |p{3.4cm}||p{1cm}|p{1.9cm}|p{1.3cm}|p{3.4cm}|p{1.6cm}| }
     \hline
     Algorithms & Greedy & Plain MCTS & Minimax & MCTS-minimax hybrid & AlphaZero\\
     \hline
     \hline
     Greedy               & -          & 10 / 0 / 0 & 10 / 0 / 0 & 10 / 0 / 0 & 10 / 0 / 0\\
     Plain MCTS           & -          & -          & 2 / 8 / 0  & 2 / 8 / 0  & 7 / 1/ 2\\
     Minimax              & -          & -          & -          & 6 / 2 / 2  & 10 / 0 / 0\\
     MCTS-minimax hybrid  & -          & -          & -          & -          & 9 / 0 / 1\\
 \hline
\end{tabular}
\end{table}

Table \ref{comp_table} compares the performance of the various agents implemented. Each agent plays a total of 10 games against every other agent, and the score is calculated as follows:\\
For comparing algorithm $i$ in row $i$ against algorithm $j$ in column $j$, the first number is the number of times $j$ wins, the second is the number of times $i$ wins (or $j$ looses) and the third number is the number of draws.

\begin{table}[h!]
  \caption{Kaggle score of agents}
  \label{score_table}
  \centering
  \begin{tabular}{ |p{3.4cm}||p{1cm}| }
     \hline
     Algorithms & Score\\
     \hline
     \hline
     Greedy               & 268\\
     Plain MCTS           & 1075\\
     Minimax              & 873\\
     MCTS-minimax hybrid  & 980\\
     AlphaZero            & \textbf{1282}\\
 \hline
\end{tabular}
\end{table}

Table \ref{score_table} lists the scores of our submissions of the implemented algorithms on Kaggle competition for ConnectX. The scoring system in Kaggle works as follows: each submission is initialised with a fixed score of 600, then the submissions are played against one another and the score of the winning submission increases while that of the losing submission decreases (proportional to the difference in their 'skill'). In case of a draw the score of the submission with the lower score increases while the submission with the higher score loses some points.

The final scores of all our submissions have been recorded in table \ref{score_table}. All the submissions except the greedy algorithm have scored over 600, at which they were initialized. These scores are also consistent with the performance we get when we play the agents against one another. The competition also maintains a leaderboard which ranks a user according to their best submission. AlphaZero, which is our best agent, has landed us at the 9th position on the leaderboard out of 225 teams with around 700 submissions as on May 5, 2022. This is much better than our initial goal of being in the Top-25, and our revised goal after the mid-stage presentation of being in the Top-15.

As is evident from table \ref{comp_table} and table \ref{score_table}, the plain MCTS agent did better than the hybrid MCTS agent with minimax based rollouts. The reason for this can be attributed to the fact that the hybrid agent is able to do much fewer simulations (around 500 per move) as compared to the plain MCTS agent (20000 simulations per move) on account of the computation cost of minimax searches for every rollout. However, since each simulation for the hybrid agent is much more efficient than that of the plain MCTS agent, the hybrid agent does not perform significantly worse (inspite of the severe shortage of iterations that MCTS requires to converge).

The results that we have obtained as a part of this project show how various algorithms play out when deployed in a simple-ruled yet difficult-to-learn game. Using our findings, we can claim that certain algorithms are better than others under the given conditions. Games are used to test algorithms since they are able to mimic specific real world game-theory problems in a compact and specific way. Hence, these results also work as a proof of concept for the algorithms.

\section{Conclusion and Future Outlook}

From our results, we see how various algorithms perform when they are played against one another. This helped us to rank these algorithms. The Kaggle submission further provided us with validation for our results and provided us with a baseline to compare the different agents. Hence, as concluded in \cite{alphazero} as well, we can say that AlphaZero is the best amongst the other algorithms tried for playing ConnectX.

One possible direction of future work is improvement in the AlphaZero algorithm. We have only used a basic architecture for the neural network, and the hyperparameters haven't been finely tuned. With the availability of sufficient time and compute power, this can be done to further improve the performance of AlphaZero. Since we are still only at the 9th position on the leaderboard of the competition, there is room for improvement to be made. 

We note that all of our algorithms had very little, if at all any, dependency on domain knowledge of ConnectX (the obvious exception being the heuristic used in minimax). We believe that such general algorithms are the way to go, and the success self-play based AlphaZero further strengthens our belief. This, however, requires large amount of computational resources for game simulation and larger neural networks along with set baselines with which to compare, like Stockfish for chess and Elmo for shogi, or a competition like the one we have used for ConnectX, which are not easily available for other games.

\ack
We would like to thank Prof. Shivaram Kalyanakrishnan for his assistance throughout the project, and especially for introducing us to the wonderful field of Reinforcement Learning.

We are also grateful to Sumeet Kumar Mishra and Gagan Jain for helping us improve the quality of our submissions and providing references to useful resources and helping out in times of need.

\printbibliography

\end{document}